# SENTIMENT ANALYSIS ON YOUTUBE SMART PHONE UNBOXING VIDEO REVIEWS IN SRI LANKA

Sherina Sally [1]

[1] Lecturer, Department of Information and Communication Technology, University of Colombo, Sri Lanka

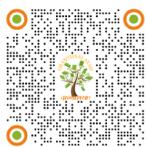
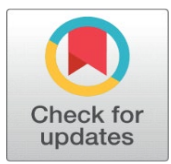

**Received** 12 October 2022
**Accepted** 12 November 2022
**Published** 30 November 2022

**Corresponding Author**
Sherina Sally, sherina.sally@gmail.com

**DOI** 10.29121/granthaalayah.v10.i11.2022.4884

**Funding:** This research received no specific grant from any funding agency in the public, commercial, or not-for-profit sectors.

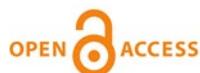

### ABSTRACT

Product-related reviews are based on users' experiences that are mostly shared on videos in YouTube. It is the second most popular website globally in 2021. People prefer to watch videos on recently released products prior to purchasing, in order to gather overall feedback and make worthy decisions. These videos are created by vloggers who are enthusiastic about technical materials and feedback is usually placed by experienced users of the product or its brand. Analyzing the sentiment of the user reviews gives useful insights into the product in general. This study is focused on three smartphone reviews, namely, Apple iPhone 13, Google Pixel 6, and Samsung Galaxy S21 which were released in 2021. VADER, which is a lexicon and rule-based sentiment analysis tool was used to classify each comment to its appropriate positive or negative orientation. All three smartphones show a positive sentiment from the users' perspective and iPhone 13 has the highest number of positive reviews. The resulting models have been tested using Naïve Bayes, Decision Tree, and Support Vector Machine. Among these three classifiers, Support Vector Machine shows higher accuracies and F1-scores.

## 1. INTRODUCTION

Since computers and other devices have become more widely used, there are no longer any limitations on how people can communicate digitally Bowman, M. (2022). Social media facilitates idea sharing, which improves interpersonal communication. Finding and exchanging product-related information is one such activity. Online reviews are the general word for product-related information provided by customers online Li and Hitt (2008). YouTube ads reached 58.9 percent of Sri Lanka's total internet user base at the beginning of 2022. In the years ahead, online reviews that incorporate videos are expected to gain more attention. According to estimates from 2022, 80% of all web traffic was made up of online videos. 90% of consumers said that watching a product review video could influence their decision to purchase it. YouTube is the largest platform for Internet videos, with the number of videos increasing by 100% annually. Inasmuch as 64% of consumers are more interested in buying a product after watching a video about it (DataReportal).

YouTube plays a vital role in providing facilities to share ideas both through video and text. It is the second largest widely used social network site and the second Most Popular Websites Worldwide (2021). YouTube is popular among vloggers who share knowledge on various products and brands. Viewers dispersed universally get the opportunity to interact with this video content by placing their comments. These reviews are useful, as they are commented on by experienced users of a specific brand or those who are willing to switch to a novel brand. Therefore, the reviews would contain a positive or negative inclination. As there can be large amounts of textual reviews for a specific video on a smartphone product, it is a challenging task to get an overview of the overall opinion. Hence, obtaining the sentiment of each comment and analysing it quantitively, would give efficacious insights into the product.

This research is conducted on three smartphones released in 2021 namely, Apple iPhone 13, Google Pixel 6, and Samsung Galaxy S21. The videos for this study were selected from a randomly chosen Sri Lankan vlogger on YouTube based on the higher number of reviews. Therefore, the comments are placed mostly by Sri Lankans. The reviews contained a number of unrelated comments which were removed during the pre-processing stage. VADER sentiment analysis tool was used to annotate the reviews as positive, negative, or neutral. Three supervised learning classifiers namely Naïve Bayes (NB), Decision Tree (DT), and Support Vector Machine (SVM) were used to test the accuracy of the data.

## 2. RELATED WORK

VADER sentiment analysis tool has been used in various domains. In research conducted to analyse the sentiment of mobile unboxing comments on YouTube, the authors have used a hybrid approach with the involvement of the Naïve Bayes algorithm and the VADER sentiment analysis technique. The hybrid approach is a combination of both the Machine Learning and lexicon-based approach. In this study V.D. (2019), the dataset consists of only comments regarding the mobile unboxing of LG G7 devices. The data is extracted using a scrapper which yielded 6248 comments and 14 metadata associated with these comments. As the authors state, only the comments have been taken into consideration for this analysis. VADER sentiment analysis is an intensity-based method of classifying text data. The comments are labelled using VADER and then classified as positive and negative. Neutral comments are eliminated from the analysis to prepare it as a binary classification which eventually will be the input to the Naïve Bayes algorithm where the data will be trained. The model achieves an accuracy of 79.78% and an F-score of 83.72%. The confusion matrix is evident that the classifier has performed well with significant results.

Handling customer support emails of corporate clients affect the reputation and customer satisfaction of the company. In many kinds of research done recently, the sentiment of the unstructured text is analysed to view for any useful insights. There is barely any research done to predict the sentiment of a not-yet-seen text such as an email response yet to be received by a customer. The authors of Borg and Boldt (2020), describe the need for such an analysis in terms of both positive and negative responses. As they claim, the customer support staff gets sufficient time to prepare for any negative reply from the customer. On the other hand, they can identify trends and plan campaigns for satisfied customers. Furthermore, it will promote the business among potential new customers through the links of satisfied customers.

Additionally, this analysis would serve as a metric for evaluating the performance of in-house and outsourced customer support personnel. It will be an indicator of how well the two groups have responded to their clients towards promoting the business venture. The data used for this analysis are customer support e-mails collected from a large Swedish telecom company over a year. VADER sentiment classifier along with a Swedish lexicon has been used to label the email corresponding to its content. In situations when there are many emails in a single thread, a compound score is calculated based on the sentiment score of all the emails in that thread. These scores are binned into five groups ranging from very negative (-1) to very positive (+1). The features of the content are extracted using the Chi-square test. The data is trained and tested using two implementations of SVM, namely SVC and LinearSVC. Additionally, these SVM models are compared against a dummy classifier which randomly classifies the classes. In order to evaluate whether the SVM models were more statistically significant than the random guesser, the Freidman test and the Nemenyi post hoc test have been used.

The importance of analysing the sentiment of Tweets in the Digital Banking sector has been highlighted by a study done by Botchway et al. (2020), for a bank in Sub-Saharan, Africa. The aim of this research was to identify the quality of the services offered to its customers on the digital platform. Instead of conventional surveys, focus groups, and interviews, a data-driven approach is insightful and robust due to the fact that the user expresses their true intentions regarding the banking application. Although this study deduces user feedback from online banking, there can be possibilities of an entirely different perspective of the services of the bank when compared with traditional banking. Data has been collected using an application named Twint. A sample of the data set has been manually labelled as positive, negative, and neutral by two independent annotators. The labelled tweets are trained using the Naïve Bayes classifier. Four sentiment lexicons namely VADER, AFINN, SentiWordNet, and TextBlob have been used to identify the sentiment polarity of the text. All the lexicons produce positive, negative, and neutral sentiments. VADER lexicon has outperformed the other lexicons used in this research. It is a lexicon with a rule-based approach that is accustomed to analysing social media text. Additionally, word clouds are generated for positive and negative sentiment words separately. These have been useful to identify the improvable and pain points of potential customer service and transactions.

The importance of opinion mining in the banking sector is highlighted in a study done by Botchway et al. (2019) for social media posts of UniCredit Bank in Europe. The authors mention about setbacks of using the traditional method of gathering opinions through polls as well as the breakthroughs of using social media platforms to obtain insights into the attitude of the public. Gauging the interests of the public is the key to the survival of business ventures in the present competitive environment. 953 English tweets associated with UniCredit Bank's official Twitter account have been used for this research. Finding any patterns from the Twitter data and the customer hashtags on products of the UniCredit bank are the main areas which this research is built upon. The VADER sentiment analysis tool has been used to determine the sentiment score of each tweet in the dataset. Based on a threshold value of previous research and the compound scores of sentiments, the tweets are labelled as positive, negative, and neutral. The positive sentiments dominate the negative and neutral categories. However, the negative sentiments have the least score. Additionally, the top ten hashtags of these tweets were identified. Results show that projects, workshops, and competitions held by the bank relating to charity projects, and finance innovations are well renowned among the customers. The high positive sentiments and the top-rated hashtags signify the degree of customer satisfaction attained by the bank.

Another study Newman and Joyner (2018) conducted to analyse student reviews on a taught course, uses the VADER sentiment tool to evaluate the teaching practices based on the student comments. Further, it determines the frequently used keywords and acquires the sentiment of comments which includes these keywords

Automated sentiment analysis was conducted using the morphological sentence pattern model, and the results showed that the accuracy of the new approach was higher (91.2%) Han and Kim (2017). Movie reviews from IMDb, Rotten Tomatoes, Metacritic, YouTube, and Twitter were used for this analysis. The system consists of four main parts: a collector, an extractor of aspects and expressions, an extractor of sentiment patterns, and an analyser of sentiment. In the first phase, the collector crawls movie reviews, tweets, and YouTube comments. The MSP model is used by the aspect-expression extractor in the second phase to locate aspects and expressions. The third step of the sentiment pattern extractor provides all feasible morphological sentence patterns. In the last stage, the sentiment analyser links patterns and emotive vocabulary to reviews of the collected data and user-generated ratings. The "Jsoup HTML parser," an open-source Java library of methods, is used by the system to extract and manipulate data that is stored in HTML. Reviews are automatically gathered by the collector. Rating systems vary from site to site. Authors can designate their reviews on Rotten Tomatoes as "Fresh" for positive or "Rotten" for unfavourable or negative. IMDb and Metacritic authors express their opinions on a scale of 1 to 10. The evaluation is more favourable the higher the number. They selected to treat ratings of 8 to 10 as positive opinions and ratings of 1 to 3 as negative opinions in order to determine the positivity of expressions. Similarly, YouTube offers APIs to gather data including video details, user profiles, and user comments. Twitter also provides tweets with the keyword included. When the target items, such as businesses, products, politicians, or movies, are searched for using certain keywords, the crawler gathers comments made on those objects. Additionally, based on user requests, it repeatedly collects the data within a predetermined timeframe. This approach's key benefit was that it could be used to reach better levels of accuracy without the need for human-coded train sets. However, there is a drawback to this strategy. The issue is that this model cannot examine the data if the pattern is not found.

On Ahok's performance as a governor, opinion mining is used Tanesab et al. (2017). The sentiment analysis is used to identify a pattern or a specific Ahok character on a sample of 1000 datasets. The Support Vector Machine is used to categorize opinions into three groups: positive, neutral, and negative. Before classifying the data, a pre-processing phase has been performed including important steps as tokenizing, cleansing, and filtering noisy characters. The Lexicon Based approach is used to calculate the percentage of class sentiment. The experiment demonstrates that the proposed approach for determining the percentage weight in this study used Lexicon Based and Confusion Matrix to determine the result of weighting percentage of analysis to SVM. The following results were discovered: accuracy 84%, precision 91%, recall 80%, TP rate 91.1, and TN rate 44.8%.

There are instances where a user's opinion contains comparisons. After watching a video comparison of two options or products, the viewer expresses his or her preference based on a few arguments. When using comparative opinion mining, the quantity of options determines the number of labels. In order to determine the opinions of the commenters on various choices, the author performed multilabel classification using the Nave Bayes machine learning technique in this study Khan et al. (2016). They made the naive assumption that words surrounding keywords associated with a specific selection would be sufficient to grasp the user's thoughts in order to lessen the computing needs. The classifier that was created using naive assumptions had a little lower performance but needed less computing power.

Using the Google API, the authors was able to extract the YouTube video comments in the csv (comma-separated values) format and, after the necessary pre-processing, determine the most up-to-date qualitative sentiment of the comments and their replies Nawaz et al. (2019). To obtain the video's recommendation label, the authors have further pursued a direct engagement on the percentage of comments with replies. A sentiment percentage of more than 50% is then determined by counting the number of positive and negative replies against each comment and computing the percentage of positive replies. In the second method, they determine the average sentiment of each comment's responses before determining the proportion of comments with positive average replies to sentiment. In order to obtain a single normalized score, they then take the arithmetic mean of the scores from the two methodologies. The method finally identifies the movie as one of the four labels based on this score: 1- not suggested 2- might be suggested 3- suggested and 4- highly suggested. On a number of YouTube comments, they test the suggested methodology, and the results are remarkably accurate when compared to human judgment. This unquestionably strengthens the YouTube search term's video selection criterion.

To analyse the expectations that social media users have for information systems (IS) products that are under development but have not yet been made available (Banerjee, Singh, Dwivedi, & Rana, 2021), the authors experiment data in Twitter related to newly released Apple and Samsung smartphones and smartwatches. The following four forthcoming IS products' tweets were discovered between 1 January 2020 and 30 September 2020: (1) Apple iPhone 12 (6,125 tweets), (2) Apple Watch 6 (553 tweets), (3) Samsung Galaxy Z Flip 2 (923 tweets), and (4) Samsung Galaxy Watch Active 3. (207 tweets). These 7,808 tweets performed sentiment analysis using the Natural Language Processing Toolkit (NLTK) (SentiWordNet). Positive tweets outnumbered negative ones in relation to every upcoming device. When it came to Apple, the most common sentiment was indifference, and when it came to Samsung, it was positive. Additionally, it was discovered that Apple had a lower percentage of tweets reflecting unfavourable sentiment than Samsung.

The author examined Twitter user behaviour and opinions using prediction-based analysis Gunasiri (2021). Its main goals were to develop election prediction-specific research and expand the corpus of resources available for text analysis. Sinhala was employed in a novel application area. To forecast the outcomes of each candidate's election, automatic labelling is applied. These anticipated outcomes were contrasted with Sri Lanka's actual 2019 presidential election outcomes. By using text pre-processing and feature extraction approaches, an appropriate model was created. To determine the best classifiers for predictive sentiment analysis in the Sinhala language, supervised learning classifiers were trained against the developed model.

As the VADER sentiment analysis tool has been used in many types of research with promising results, it was decided to apply it in this study.

## 3. METHODOLOGY
### 3.1. DATA COLLECTION

The data for this study was obtained by executing a Python script to extract the comments of three smartphone unboxing videos namely iPhone 13, Google Pixel 6, and Samsung Galaxy S21. Initially, there were 11173 comments for iPhone, 13708 comments for Google Pixel 6, and 8630 comments for Samsung Galaxy S21. The comments were downloaded in the JSON format and then, parsed into an excel worksheet using OpenRefine[1]. The varied numbers in comments collected are due to the release dates of these different smartphones.

### 3.2. DATA PREPROCESSING

The data collected contained texts, numerals, and emoticons. Most of the textual comments were in English whereas a few were in Sinhala, Russian and Chinese. For this research, only comments in English were considered. Comments which comprised only numerals were eliminated as they did not represent any meaning. By observation, it was identified that there were a number of repetitive comments and capitalized text which could be considered generated by a bot. They followed a similar pattern. Most of the capitalized texts acquired more than half of the comments' text length which denoted an unusual pattern. Subsequently, they were meaningless. These were removed from the dataset using a Python script written by matching RegEx[2]. Finally, only unique comments were considered for processing.

### 3.3. LABELLING THE DATA

The comments were labelled using the VADER (Valence Aware Dictionary and sEntiment Reasoner) tool. VADER is a lexicon and rule-based sentiment analysis tool which annotates a text in the positive, negative, or neutral orientation as well as obtains the compound score which is the aggregation of all three scores. Moreover, it handles social media text well, which heavily consists of emojis, acronyms, contractions, and slang words. The following conditions were applied to the compound scores, in order to classify them to the respective sentiment.

- 0.5 ≥ compound score ← negative

- 0.5 < compound score < 0.5 ← neutral

0.5 ≤ compound score ← positive

After the classification, only the negative and positive comments were used for the rest of the study. Table 1 denotes the composition of the comments for all three smartphones. According to the results obtained, all three smartphones have a positive sentiment in general. The positive comments on iPhone 13 and Pixel 6 are more than 2.0 times higher than the negative comments whereas, on Galaxy S21 it is only 1.6 times higher. However, iPhone 13 has received the highest positive opinion of the other two devices.

---

[1] https://openrefine.org/
[2] https://docs.python.org/3/library/re.html

**Table 1**

| Table 1 Positive and Negative Comments After Cleaning | | | |
|---|---|---|---|
| Smart Phone | Positive | Negative | Total |
| iPhone 13 | 4377 | 1527 | 5904 |
| Pixel 6 | 4165 | 1894 | 6059 |
| Galaxy S21 | 3337 | 2033 | 5370 |

### 3.4. EXTRACTING FEATURES

The comments include useful features relevant to the context. Prior to extracting the features, stop words were removed from the text and the words were lemmatized using spaCy[3] library. Afterward, features of the text were identified using the count vectorizer tool. It transforms a text into a vector, based on the frequencies of each word occurred in the entire text. All three smartphone reviews had approximately two hundred features with a min_df of 0.01 which indicates that terms appearing less than 1% in the document were ignored.

### 3.5. EVALUATION

The final stage of this research is to evaluate the reviews that were annotated using VADER. Three machine learning classifiers were used for this purpose namely Naïve Bayes, Decision Tree, and Support Vector Machine. The data were split into 80% training and 20% testing data.

### 4. RESULTS AND DISCUSSION

The confusion matrix for Naïve Bayes, Decision Tree, and Support Vector Machine is illustrated in Table 2, Table 3, and Table 4 .

According to the results in Table 2, for Samsung Galaxy S21, 211 True Negatives and 579 True Positives were identified which sums up to 790 correctly predicted comments out of a total of 1074 comments. For Google Pixel 6, 144 True Negatives and 764 True Positives were identified which sums up to 908 correctly predicted comments out of a total of 1212 comments. For iPhone 13, 41 True Negatives and 874 True Positives were identified which sums up to 915 correctly predicted comments out of a total of 1181 comments.

**Table 2**

| Table 2 Confusion Matrix of Naïve Bayes | | | | | |
|---|---|---|---|---|---|
| | | Predicted | | | Samsung Galaxy S21 |
| | | Negative | Positive | All | |
| Actual | Negative | 211 | 209 | 420 | |
| | Positive | 75 | 579 | 654 | |
| | All | 286 | 788 | 1074 | |
| | | Predicted | | | Google Pixel 6 |
| | | Negative | Positive | All | |

---

[3] https://spacy.io/

| | | Negative | Positive | All | |
|---|---|---|---|---|---|
| Actual | Negative | 144 | 221 | 365 | |
| | Positive | 83 | 764 | 847 | |
| | All | 227 | 985 | 1212 | |
| | | Predicted | | | iPhone 13 |
| | | Negative | Positive | All | |
| Actual | Negative | 41 | 258 | 299 | |
| | Positive | 8 | 874 | 882 | |
| | All | 49 | 1132 | 1181 | |

On the other hand, according to the results in Table 3, Samsung Galaxy S21 produced, 228 True Negatives and 496 True Positives which sums up to 724 correctly predicted comments out of a total of 1074 comments. For Google Pixel 6, 182 True Negatives and 650 True Positives were identified which sums up to 832 correctly predicted comments out of a total of 1212 comments. For iPhone 13, 144 True Negatives and 729 True Positives were identified which sums up to 873 correctly predicted comments out of a total of 1181 comments.

**Table 3**

| **Table 3 Confusion Matrix of Decision Trees** | | | | | |
|---|---|---|---|---|---|
| | | **Predicted** | | | Samsung Galaxy S21 |
| | | **Negative** | **Positive** | **All** | |
| Actual | Negative | 228 | 192 | 420 | |
| | Positive | 158 | 496 | 654 | |
| | All | 386 | 688 | 1074 | |
| | | Predicted | | | Google Pixel 6 |
| | | Negative | Positive | All | |
| Actual | Negative | 182 | 183 | 365 | |
| | Positive | 197 | 650 | 847 | |
| | All | 379 | 833 | 1212 | |
| | | Predicted | | | iPhone 13 |
| | | Negative | Positive | All | |
| Actual | Negative | 144 | 155 | 299 | |
| | Positive | 153 | 729 | 882 | |
| | All | 297 | 884 | 1181 | |

Similarly, as in Table 4, Samsung Galaxy S21 has 255 True Negatives and 541 True Positives which sums up to 796 correctly predicted comments out of a total of 1074 comments. For Google Pixel 6, 138 True Negatives and 789 True Positives were identified which sums up to 927 correctly predicted comments out of a total of 1212 comments. For iPhone 13, 63 True Negatives and 859 True Positives were identified which sums up to 922 correctly predicted comments out of a total of 1181 comments.

**Table 4**

| Table 4 Confusion Matrix of Support Vector Machine | | | | | |
|---|---|---|---|---|---|
| | | **Predicted** | | | Samsung Galaxy S21 |
| | | **Negative** | **Positive** | **All** | |
| Actual | Negative | 255 | 165 | 420 | |
| | Positive | 113 | 541 | 654 | |
| | All | 368 | 706 | 1074 | |
| | | Predicted | | | Google Pixel 6 |
| | | Negative | Positive | All | |
| Actual | Negative | 138 | 227 | 365 | |
| | Positive | 58 | 789 | 847 | |
| | All | 196 | 1016 | 1212 | |
| | | Predicted | | | iPhone 13 |
| | | Negative | Positive | All | |
| Actual | Negative | 63 | 236 | 299 | |
| | Positive | 23 | 859 | 882 | |
| | All | 86 | 1095 | 1181 | |

By using the macro-average on each precision, recall, and F1-score, the accuracies have been generated for each smartphone under each classifier. The accuracy is higher in Support Vector Machine compared to the other two classifiers according to the results generated in Table 5, Table 6, and Table 7. The least accuracies are observed in the Decision Tree. Naïve Bayes has a similar value as the Support Vector Machine in terms of accuracy. The F1-score follows the same pattern.

**Table 5**

| Table 5 Classification Report for Samsung S21 | | | | |
|---|---|---|---|---|
| | **Precision** | **Recall** | **F1-Score** | **Accuracy** |
| NB | 0.74 | 0.69 | 0.70 | 0.74 |
| DT | 0.66 | 0.65 | 0.65 | 0.67 |
| SVM | 0.73 | 0.72 | 0.72 | 0.74 |

**Table 6**

| Table 6 Classification Report for Google Pixel6 | | | | |
|---|---|---|---|---|
| | Precision | Recall | F1-Score | Accuracy |
| NB | 0.70 | 0.65 | 0.66 | 0.75 |
| DT | 0.63 | 0.63 | 0.63 | 0.69 |
| SVM | 0.74 | 0.65 | 0.67 | 0.76 |

**Table 7**

| Table 7 Classification Report for iphone13 | | | | |
|---|---|---|---|---|
| | Precision | Recall | F1-Score | Accuracy |
| NB | 0.80 | 0.56 | 0.55 | 0.77 |
| DT | 0.65 | 0.65 | 0.65 | 0.74 |
| SVM | 0.76 | 0.59 | 0.60 | 0.78 |

## 5. CONCLUSION

This research was conducted to compare the opinion orientation of three smartphones released in 2021 by three popular mobile companies namely Apple Inc., Samsung Electronics Co., Ltd, and Google LLC. People are naturally inclined to know the views of others, on a product, before purchasing. The reason is that most probably the reviews are placed by experienced or enthusiastic users of a product or brand. These users can have either positive or negative thoughts based on their experience. This research was conducted to find insights into such opinions of the users on the smartphones chosen. Comments that were identified as generated by bots and were meaningless were removed. Furthermore, comments with only a numeric value, more capitalized characters than half of the total characters in a comment, and non-English text were not considered for the analysis.

VADER sentiment tool was used to annotate each comment as positive, negative, or neutral. The neutral comments were removed from the analysis. It was observed that the positive comments are higher than the negative comments for all smartphones. Hence, it could be derived that, all three smartphones have a positive opinion from their' users' perspective and the iPhone 13 has received the highest positive comments among the three. Three machine learning classifiers were used to test the annotated text. Support Vector Machine outperforms the other two classifiers namely Naïve Bayes and Decision Tree algorithm.

### CONFLICT OF INTERESTS
None.

### ACKNOWLEDGMENTS
None.

### REFERENCES

Banerjee, S., Singh, J. P., Dwivedi, Y. K., & Rana, N. P. (2021). Social Media Analytics for End-Users' Expectation Management in Information Systems

Development Projects. Information Technology & People, 34(6), 1600-1614. https://doi.org/10.1108/ITP-10-2020-0706

Borg, A., & Boldt, M. (2020). Using VADER sentiment and SVM for Predicting Customer Response Sentiment. Expert Systems with Applications, 162. https://doi.org/10.1016/j.eswa.2020.113746

Botchway, R. K., Jibril, A. B., & Komínková, Z. (2020). Deductions from a Sub-Saharan African Bank's Tweets: A Sentiment Analysis Approach. Cogent Economics & Finance, 8(1). https://doi.org/10.1080/23322039.2020.1776006

Botchway, R. K., Jibril, A. B., Kwarteng, M. A., Chovancova, M., & Oplatková, Z. K. (2019). A Review of Social Media Posts from Unicredit Bank in Europe: A Sentiment Analysis Approach. Proceedings of the 3rd International Conference on Business and Information Management, (pp. 74-79). New York, USA. https://doi.org/10.1145/3361785.3361814

Bowman, M. (2022). "Video Marketing : the Future Content Marketing." Forbes.

DataReportal. (2022).

Gunasiri, W. (2021). Sentiment Analysis of Tweets to predict Sri Lankan Election Results using Supervised Learning Techniques. University of Colombo School of Computing.

Han, Y., & Kim, K. K. (2017). Sentiment Analysis on Social Media Using Morphological Sentence Pattern Model. 2017 IEEE 15th International Conference on Software Engineering Research, Management and Applications (SERA). London, UK: IEEE. https://doi.org/10.1109/SERA.2017.7965710

Khan, A. U., Khan, M., & Khan, M. B. (2016). Naive Multi-label Classification of YouTube Comments Using Comparative Opinion Mining. Procedia Computer Science, 82, 57-64. https://doi.org/10.1016/j.procs.2016.04.009

Li, X., & Hitt, L. M. (2008). Self-selection and Information Role of Online Product Reviews. Information Systems Research, 19(4), 456-474. https://doi.org/10.1287/isre.1070.0154

Most Popular Websites Worldwide (2021). As of June, By Total Visits. (2021, 12 19).

Nawaz, S., Rizwan, M., & Rafiq, M. (2019). Recommendation Of Effectiveness Of Youtube Video Contents By Qualitative Sentiment Analysis Of Its Comments And Replies. Pakistan Journal of Science, 71(4).

Newman, H., & Joyner, D. (2018). Sentiment Analysis of Student Evaluations of Teaching. International Conference on Artificial Intelligence in Education, 246-250. https://doi.org/10.1007/978-3-319-93846-2_45

Tanesab, F. I., Sembiring, I., & Purnomo, H. D. (2017). Sentiment Analysis Model Based OnYoutube Comment Using Support Vector Machine. International Journal of Computer Science and Software Engineering, 6(8), 180-185.

V.D., C. (2019). Hybrid Approach: Naive Bayes And Sentiment VADER for Analyzing Sentiment of Mobile Unboxing Video Comments. International Journal of Electrical and Computer Engineering, 4452-4459. http://doi.org/10.11591/ijece.v9i5.pp4452-4459